\documentclass[letterpaper, 10 pt, journal]{IEEEtran}
\IEEEoverridecommandlockouts                          
\usepackage{graphics} 
\usepackage{epsfig}
\usepackage{times}
\usepackage{amsmath}
\usepackage{amssymb}
\usepackage{threeparttable}
\usepackage{booktabs}
\usepackage{soul}
\usepackage{subfigure}
\usepackage{xcolor}
\usepackage[hidelinks]{hyperref}
\usepackage{algorithm}
\usepackage{algorithmic,comment}
\usepackage{tcolorbox}

\title{\LARGE \bf
PrediFlow: A Flow-Based Prediction-Refinement Framework for Real-Time Human Motion Prediction in Human-Robot Collaboration
}

\author{Sibo Tian$^{1}$, Minghui Zheng$^{1,*}$, and Xiao Liang$^{2,*}$
\thanks{This work was supported by the USA National Science Foundation under Grant No. 2026533/2422826. The authors confirm that all human/animal subject research procedures and protocols are approved by the review board. Portions of this research were conducted with the advanced computing resources provided by Texas A\&M High Performance Research Computing.}
\thanks{$^{1}$ Sibo Tian and Minghui Zheng are with the J. Mike Walker '66 Department of Mechanical Engineering, Texas A\&M University, College Station, TX 77843, USA. {\tt\small Emails: {sibotian, mhzheng}@tamu.edu.}}
\thanks{$^{2}$ Xiao Liang is with the Zachry Department of Civil and Environmental Engineering, Texas A\&M University, College Station, TX 77843, USA. {\tt\small Email: xliang@tamu.edu.}}
\thanks{$^{*}$ Corresponding Authors.}}

\begin{document}

\maketitle
\thispagestyle{empty}
\pagestyle{empty}

\begin{abstract}
Stochastic human motion prediction is critical for safe and effective human-robot collaboration (HRC) in industrial remanufacturing, as it captures human motion uncertainties and multi-modal behaviors that deterministic methods cannot handle. While earlier works emphasize highly diverse predictions, they often generate unrealistic human motions. More recent methods focus on accuracy and real-time performance, yet there remains potential to improve prediction quality further without exceeding time budgets. Additionally, current research on stochastic human motion prediction in HRC typically considers human motion in isolation, neglecting the influence of robot motion on human behavior. To address these research gaps and enable real-time, realistic, and interaction-aware human motion prediction, we propose a novel prediction-refinement framework that integrates both human and robot observed motion to refine the initial predictions produced by a pretrained state-of-the-art predictor. The refinement module employs a Flow Matching structure to account for uncertainty. Experimental studies on the HRC desktop disassembly dataset demonstrate that our method significantly improves prediction accuracy while preserving the uncertainties and multi-modalities of human motion. Moreover, the total inference time of the proposed framework remains within the time budget, highlighting the effectiveness and practicality of our approach.
\end{abstract}

\def\abstractname{Note to Practitioners}
\begin{abstract}
This paper was motivated by the challenge of predicting reasonably diverse human behavior in human-robot collaborative disassembly, but the same ideas extend to other multi-agent collaborative applications, where an agent needs to anticipate others' future states in order to proactively plan its own actions for the safety and efficiency of the overall system. State-of-the-art work trains a single-inference-step diffusion model through knowledge distillation to improve inference speed while maintaining prediction fidelity. In this paper, we propose that even high-quality prediction results can benefit from a refinement stage. Additionally, human motion is strongly influenced by robots during collaborative tasks, but such interactions are often overlooked in previous works. To this end, we design a Flow Matching-based interaction-aware motion refiner neural network that accounts for the influence of other moving agents, such as a robot in our case, on the behavior of the agent of interest. We evaluated our method on an existing human-robot collaborative disassembly dataset and compared the performance with other works. The results show that our prediction-refinement framework can significantly improve prediction accuracy while keeping the total inference time within the real-time prediction budget. Given the limited availability of collaborative remanufacturing human motion datasets, future research will focus on designing more complex, diverse, and interaction-rich HRC scenarios, building a comprehensive HRC dataset, and further evaluating the proposed method.
\end{abstract}


\section{Introduction}
Human-robot collaboration (HRC) is becoming increasingly important in industrial remanufacturing, particularly in disassembly processes, where humans and robots work side by side, leveraging their respective strengths to efficiently break down products for component reuse, recycling, and waste reduction, contributing to both economic and environmental benefits \cite{lee2024review}. In contrast to industrial product assembly, which follows a specific and well-defined installation sequence, disassembly in remanufacturing is characterized by significant uncertainty. This uncertainty stems from factors such as the varying quality of end-of-life products, the unpredictable condition of components, and the absence of standardized procedures.  In this context, human workers, with their excellent decision-making skills and adaptability, may adjust their actions frequently based on the specific task, environment, or unexpected circumstances. This variability introduces a significant challenge for robots to seamlessly collaborate with humans while preventing potential safety hazards. Therefore, advanced human motion prediction algorithms, which enable robots to anticipate human behavior, plan their movements proactively, and adapt in real time, are essential in the HRC remanufacturing system.

\begin{figure}
    \begin{center}
        \includegraphics[width=1.0\columnwidth]{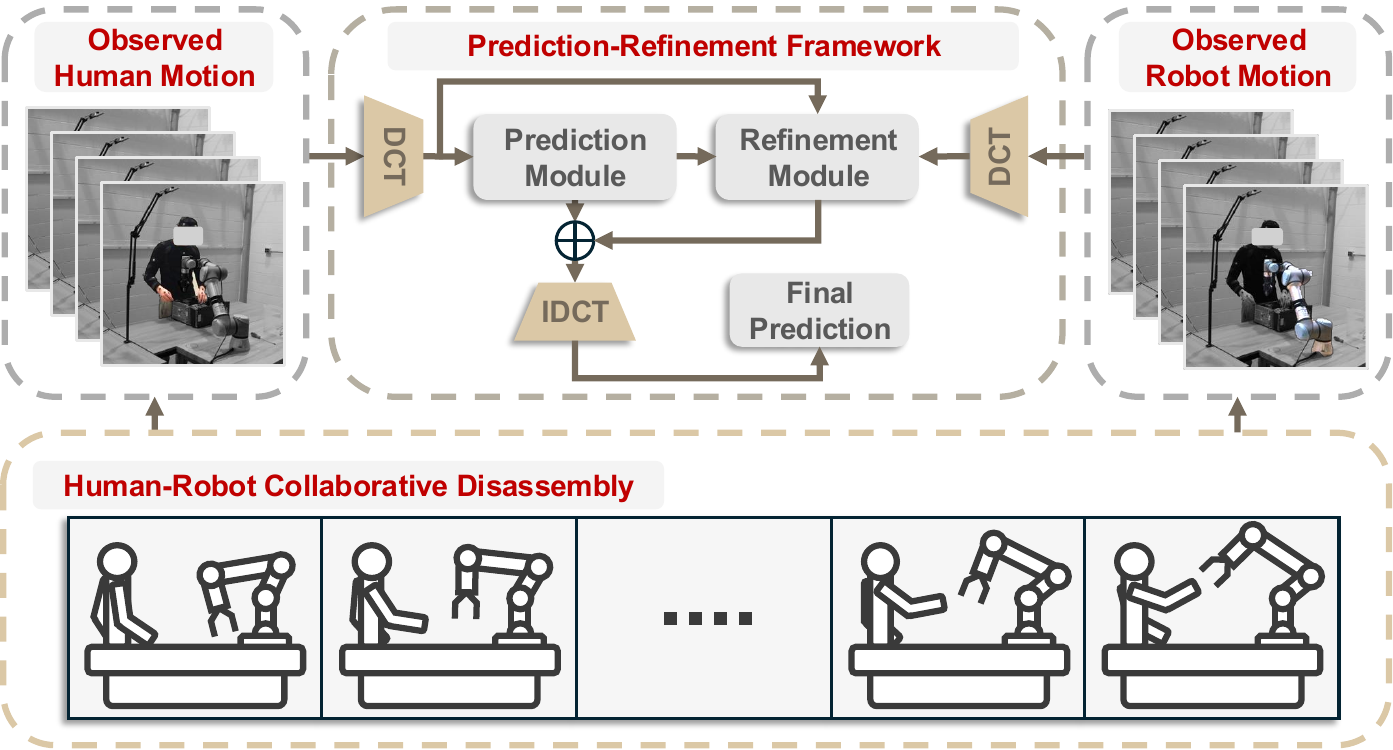}
    \vspace{-0.2in}
    \caption{An overview of the Prediction-Refinement framework: The prediction module generates multiple coarse predictions based on observed human motion, and the refinement module improves them using both observed human and robot motion.}
    \label{hrc}
    \end{center}
\end{figure}

Over the last decade, researchers have made significant advancements in human motion prediction using deep learning techniques. Models such as Recurrent Neural Networks (RNNs) \cite{noppeney2024human, zhang2022pimnet} and Graph Convolutional Networks (GCNs) \cite{du2025skeleton} have been proposed to predict human motion in deterministic manner. Despite the promising results demonstrated by these works, they ignore the inherent uncertainty and multi-modality of human motion. For safety-critical industrial applications such as HRC disassembly, where humans and robots operate in close proximity, the ability to capture a diverse range of possible future human movements is essential. In such cases, a single deterministic prediction may not accurately represent human future motion. Consequently, robots may struggle to respond appropriately to human actions, increasing the risk of miscoordination or potential hazards.

To tackle this issue, Variational Autoencoders (VAEs) \cite{yuan2020dlow} and Generative Adversarial Networks (GANs) \cite{gurumurthy2017deligan} have been utilized to learn distributions over possible future motions. Recent works have also adopted diffusion models for stochastic prediction tasks \cite{tian2024transfusion, wei2023human, chen2023humanmac}, as they generally exhibit superior generation performance compared to VAEs and GANs with a simple training procedure, though their iterative denoising steps lead to high inference latency. A state-of-the-art one-step diffusion model \cite{tian2024bayesian} targets real-time stochastic human motion prediction by distilling knowledge from a pretrained diffusion model \cite{tian2024transfusion} into a lightweight multilayer perceptron (MLP). This approach reduces inference time by 98\% and predicts motions in around 0.01 seconds without sacrificing accuracy. The authors also design an HRC desktop disassembly experiment but ignore human-robot interaction during prediction, treating human motion in isolation.

Considering the state-of-the-art real-time stochastic motion predictor \cite{tian2024bayesian} remains a large amount of prediction time budget, to further enhance prediction quality, we propose a novel interaction-aware refinement model, based on Flow Matching \cite{lipman2023flow}, that accounts for the influence of robot motion on human movements. Building on this, we formulate a prediction-refinement framework for real-time human motion prediction in HRC, following a coarse-to-fine strategy, as illustrated in Figure \ref{hrc}. Specifically, the pretrained prediction module generates multiple initial predictions based on the observed human motion. Each prediction result is then used as a condition to learn and predict the residual that represents the difference between the ground truth and the initial prediction through the refinement module. For each initial prediction, multiple refinement residuals are generated simultaneously to account for uncertainty. The final predictions are then obtained by adding the predicted residuals to the initial predictions.

In contrast to prior motion prediction and refinement approaches, Flow Matching offers distinct advantages for human motion prediction tasks. First, it is a generative model capable of capturing uncertainty arising from both the inherent variability in human motion data and the model's limited generalization to unseen inputs. By accounting for both data uncertainty and model uncertainty, Flow Matching enhances robustness and improves the reliability of predictions in safety-critical industrial applications, such as HRC end-of-life product recycling. Second, unlike diffusion models, Flow Matching encourages learning an optimal transport path, promoting a straight-line trajectory from the normal distribution to the data distribution. Since inference time is critical in human motion prediction tasks, diffusion models typically require distillation to enable one-step generation. However, this distillation process can degrade generation quality and incur additional computational costs. In contrast, Flow Matching inherently learns the straight-line transport path, enabling efficient one-step generation without the need for distillation. Overall, our contributions can be summarized as follows:

\begin{itemize}
\item We formulate the problem of real-time human motion prediction in HRC as a prediction-refinement process. The refiner module is based on the Flow Matching framework, enabling it to effectively capture uncertainties arising from both data variability and model limitations, thereby enhancing the overall refinement performance.
\item We design a new refiner architecture that accounts for the influence of robot motion on human behavior by encoding observed robot motion as a conditional input for modulation. By learning human-robot interactions, the refiner effectively improves prediction accuracy.
\item We evaluate the proposed framework on the HRC desktop disassembly dataset to emulate real-world industrial applications. The results demonstrate that our method outperforms other baselines while keeping the overall prediction time within the time budget.
\end{itemize}

\section{Related Work}

\subsection{Stochastic Human Motion Prediction}

Early in the research, most stochastic human motion prediction methods favor diversity over accuracy. For example, DLow \cite{yuan2020dlow} employs a conditional VAE framework and utilizes learnable mapping functions to transform the random variable into diverse latent codes, explicitly increasing variation in prediction results. Similarly, an anchor-based sampling method is proposed in \cite{xu2022diverse}. The authors disentangle latent codes into deterministic anchors representing motion modes and random variables representing the variation within a specific motion type. By increasing the number of anchors, the predicted motion spans a broader range of motion behaviors, enhancing prediction diversity. Additionally, diversity-promoting loss is incorporated in \cite{wei2023human} to increase the variation of predicted human motions. These diversity-enhancing techniques are originally proposed to address the mode collapse problem in vanilla generative models, where samples are predominantly drawn from the major modes while neglecting other possible minor modes. However, such blind race of increased diversity often causes motions to deviate from the ground truth early in the prediction horizon, resulting in many predictions that are neither realistic nor plausible given the context of the observed motion. Moreover, these works focus only on the error of a single prediction that is closest to the ground truth among all predictions generated, i.e., the best-of-many error \cite{yuan2020dlow}, while ignoring the quality of the remaining predictions. Consequently, a model could produce one accurate prediction while generating many unrealistic or physically implausible motions, yet still achieve a low best-of-many error.

Recently, a diffusion-based human motion prediction framework called TransFusion \cite{tian2024transfusion} emphasizes the significance of overall prediction accuracy. The authors argue that diffusion models can faithfully capture a broader range of the training distribution than VAEs and GANs, enabling the generation of diverse yet realistic trajectories that address uncertainty and multi-modality. Accordingly, TransFusion avoids explicit diversity-promoting techniques, instead learning the true distribution from data, and achieves state-of-the-art accuracy across best-of-many, median-of-many, and worst-of-many cases.

While achieving good accuracy, most stochastic human motion prediction works, especially diffusion-based methods, overlook a critical factor that determines whether the model can be deployed in real-world scenarios: inference time. Diffusion models \cite{tian2024transfusion, wei2023human, chen2023humanmac} typically rely on iterative denoising to generate clean predictions from pure noise, leading to significant latency. To address this limitation, SwiftDiff \cite{tian2024bayesian} employs a two-stage knowledge distillation process and Bayesian optimization to train a one-step MLP-based diffusion model for motion prediction. While significantly improving inference time by 98\% and enabling real-time prediction, SwiftDiff does not compromise prediction quality and still maintains state-of-the-art overall accuracy.

\subsection{Human Motion Prediction Refinement}

The refinement model improves the framework's accuracy in a coarse-to-fine manner by analyzing the discrepancy between the initial prediction and the ground truth. Although its application in human pose estimation from images or videos has been well studied \cite{fieraru2018learning, nie2018hierarchical, moon2019posefix}, refinement in 3D human motion prediction remains underexplored. To the best of the authors' knowledge, only two previous works have incorporated a refinement module in the 3D human motion prediction system \cite{wei2023human, chao2020adversarial}. Both works leverage GCN as refiner to correct coarse predictions. MotionDiff \cite{wei2023human} proposes a diffusion-refinement structure and introduces a loss term that penalizes the difference between the final prediction and the ground truth. In addition to the refiner, ARNet \cite{chao2020adversarial} also introduces adversarial learning-based error distribution augmentation to improve generalization performance. However, both of these works rely on a deterministic refiner, which fails to capture uncertainty in the refinement stage. Furthermore, these works add a refinement module in the prediction framework without carefully validating its feasibility, i.e., without comparing the time required for the coarse prediction with the available time budget, leading to an additional increase in inference latency.

\subsection{Interaction-aware Human Motion Prediction}

Context-aware motion prediction has been a key research focus in vehicle trajectory prediction \cite{radwan2020multimodal, wang2025cmp, han2023interaction}, where environmental information and the trajectories of surrounding agents are used to predict an individual agent's 2D trajectory. Despite recent advancements in 3D human motion prediction algorithms, most existing approaches treat humans in isolation, overlooking the inherent influence of interactions with surrounding objects and agents. Several works \cite{adeli2020socially, corona2020context, xia2025enhancing, zhang2024multi} have begun to incorporate such contextual information to explore the interaction-aware 3D human motion prediction. Multi-person motion prediction is considered in \cite{adeli2020socially}. Each individual’s motion is encoded independently and aggregated via pooling to obtain social features, which are then concatenated with contextual image features to predict socially and contextually aware future motion. A semantic graph model is proposed in \cite{corona2020context}, where nodes represent humans and objects, and edges capture their interactions. Combined with an RNN, it improves prediction accuracy and contextual coherence. Human-object interaction is considered in the motion prediction task in \cite{xia2025enhancing}, where both human and object future motions are predicted through a shared attention mechanism to enhance prediction accuracy. Similarly, HRC is considered in \cite{zhang2024multi}, where the authors present a prediction model that takes both human and robot motion as inputs to enhance prediction performance. While previous works have made progress in incorporating contextual information, effectively predicting interaction-aware human motion in complex scenarios, such as human-robot collaborative remanufacturing, remains challenging.

\section{Methodology}

\subsection{Problem Formulation and Notation}

Human motion prediction aims to forecast a sequence of future human motion based on past information. We denote the observed human motion history as $ X_o^h = [x_1^h, x_2^h, \dots, x_T^h] \in \mathbb{R}^{3J \times T} $, where $ x_T^h \in \mathbb{R}^{3J} $ represents the 3D coordinates of the human skeleton at time frame $ T $, and $ J $ is the number of human body joints. Similarly, the ground truth future human motion sequence is denoted as $ X_p^h = [x_{T+1}^h, x_{T+2}^h, \dots, x_{T+F}^h] \in \mathbb{R}^{3J \times F} $. The robot observed motion is defined as $ X_o^r = [x_1^r, x_2^r, \dots, x_T^r] \in \mathbb{R}^{3K \times T} $, where $ x_T^r \in \mathbb{R}^{3K} $ represents the 3D coordinates of robot joints at time frame $ T $, and $ K $ is the number of robot joints. The values of $ x_T^r $ are computed using forward kinematics based on robot joint values at each time step. Thus, the human motion prediction task in HRC is defined as minimizing the difference between $ X_p^h $ and $ \hat{X}_p^h $, where $ \hat{X}_p^h = \mathcal{F}(X_o^h, X_o^r) $ is the predicted human future motion, $X_o^h$ and $ X_o^r$ are expressed in the same coordinate system with the origin located at the robot base, and $ \mathcal{F}(\cdot) $ is the prediction model.

We predict human motion in the frequency domain, which effectively encodes temporal information and typically provides better performance, as presented in \cite{mao2019learning}. Specifically, we denote the complete human motion sequence, including both history and future, as $ X^h = [X_o^h, X_p^h] \in \mathbb{R}^{3J \times (T+F)} $. We use $ Y $ to represent the frequency coefficients obtained by applying the discrete cosine transform (DCT) to $ X $, i.e., $ Y = DCT(X) $, $ Y \in \mathbb{R}^{3J \times \tau}$ where $\tau$ denotes the number of frequency terms retained, and the inverse process is $ X = IDCT(Y) $. This frequency transformation is applied to the complete motion sequence to improve the consistency between the observed history and the predicted future motion.

In this work, we define $ \mathcal{F}(\cdot) $ as a prediction-refinement framework. First, the prediction module generates an initial prediction $ \hat{Y}_{init}^h \in  \mathbb{R}^{3J \times \tau }$. Conditioned on this, the refinement module predicts the corresponding residual $\Delta\hat{Y}^h \in \mathbb{R}^{3J \times \tau }$. Finally, the predicted human motion can be calculated as:
\begin{equation}
    \hat{X}^h = IDCT(\hat{Y}_{init}^h + \Delta\hat{Y}^h)
\end{equation}
and $\hat{X}^h_p$ can be indexed from $\hat{X}^h$.

\subsection{Flow Matching for Prediction Refinement}

\begin{figure*}
    \begin{center}
    \includegraphics[width=2.0\columnwidth]{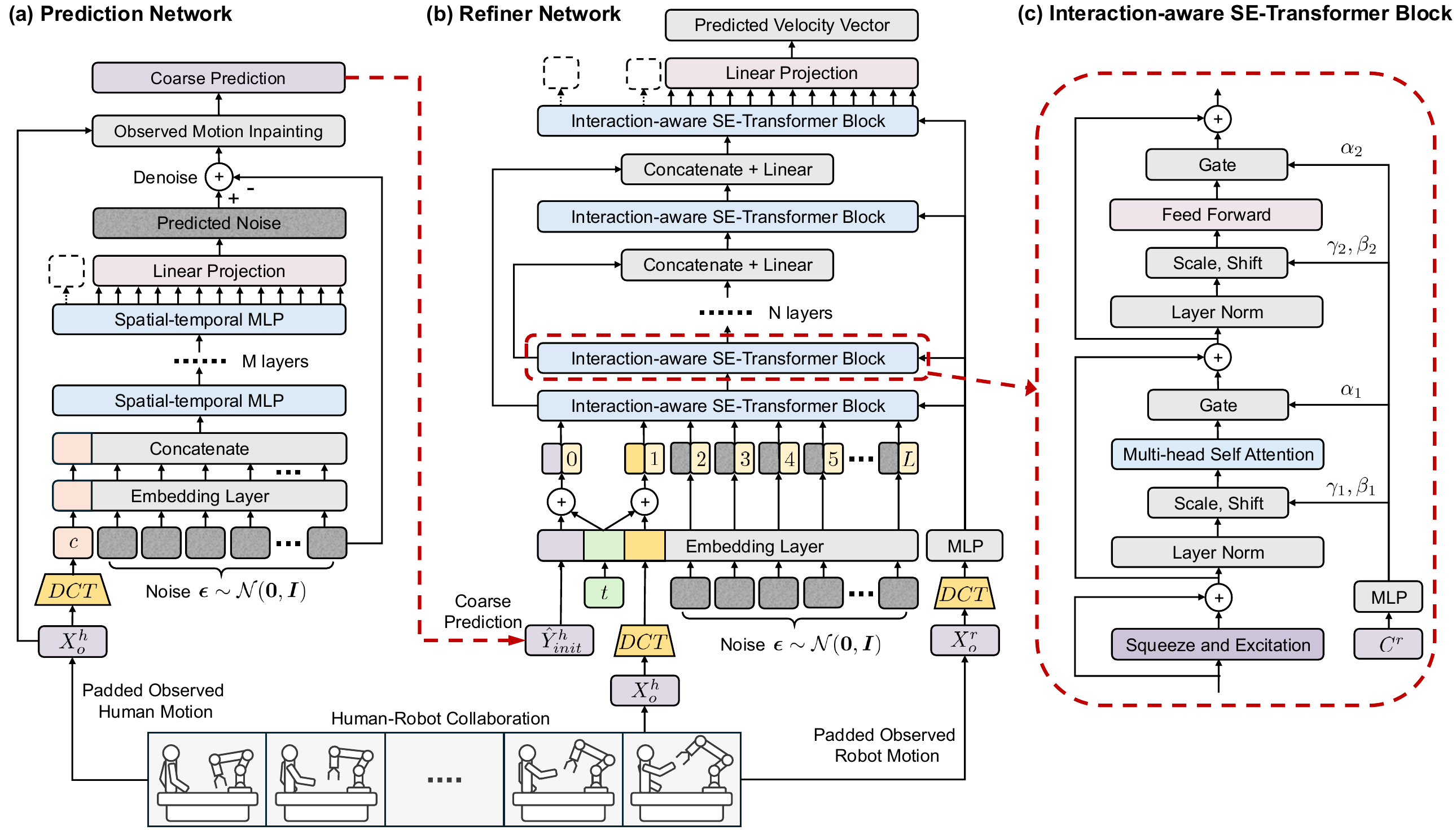}
    \caption{Neural Network Architecture of PrediFlow. Figures a and b show the detailed structure of prediction and refinement module in our prediction-refinement framework. Figure c shows the structure interaction-aware SE-Transformer block used in the refiner.}
    \label{architecture}
    \end{center}
\end{figure*}

The goal of the refinement module is to compensate for errors in the coarse prediction by learning the residual $\Delta Y^h$, given the initial prediction $\hat{Y}_{init}^h$ and conditions such as the observed human motion $X_o^h$ and robot motion $X_o^r$. To account for the uncertainty in $\Delta Y^h$ and enhance the performance, we construct our refinement module using a flow-based model.

Flow Matching \cite{lipman2023flow} is a recently developed generative modeling framework that provides a variant to diffusion models for learning complex probability distributions. It builds a probability path $ ( p_t )_{0 \leq t \leq 1} $ from a known source distribution $ p_0 = p $, usually a normal distribution, to the target data distribution $ p_1 = q $. Each $ p_t $ is a distribution over $ \mathbb{R}^d $. This path is parameterized by a velocity field that describes how sample points should move over time. Formally, an ordinary differential equation (ODE) is defined by a time-dependent velocity field $ u_t $, which is implemented using a neural network, and a time-dependent flow $ \psi_t $ where $ \psi_0(x) = x $:
\begin{equation}
    \frac{d}{dt} \psi_t(x) = u_t(\psi_t(x)).
\end{equation}
If the flow satisfies that for samples drawn from the source distribution $ \mathbf{x}_0 \sim p_0$, $ \mathbf{x}_t = \psi_t(\mathbf{x}_0) \sim p_t$, then the corresponding velocity field generates the probability path. By solving the above ODE from $ t=0 $ to $ t=1 $, we can generate samples $\mathbf{x}_1$ resembling the target data distribution $q$. Therefore, there are two main steps for training a Flow Matching model: defining a probability path $p_t$ that interpolates between $p$ and $q$, and training a neural network $ u_t^\theta $ to generate the probability path.

We use the normal distribution as the source distribution in the framework, and construct the probability path $p_t$ using the optimal transport path \cite{lipman2023flow} which promotes a straight trajectory, enabling efficient one-step generation without the need for distillation. Thus, we can get the a set of random variable $\mathbf{x}_t \sim p_t$ by taking the linear interpolation of $\mathbf{x}_0 \sim p$ and $\mathbf{x}_1 \sim q$, i.e.,
\begin{equation}
    \mathbf{x}_t = t\mathbf{x}_1 + (1-t)\mathbf{x}_0. 
\end{equation}

After defining the probability path, the velocity field neural network can be trained using the loss function defined in \cite{lipman2023flow}:
\begin{equation}
    \mathcal{L}_{\text{FM}}(\theta) = \mathbb{E}_{t, \mathbf{x}_0, \mathbf{x}_1} [ \| u_t^\theta (\mathbf{x}_t) - (\mathbf{x}_1 - \mathbf{x}_0) \|^2 ]
\end{equation}
where $t \sim \mathcal{U}[0,1]$, $\mathbf{x}_0 \sim \mathcal{N}(0,I)$ and $\mathbf{x}_1 \sim q$.

As for the residual learning for refinement purpose in our case, we have the target residual calculated by
\begin{equation}
    \Delta Y^h = Y^h - \hat{Y}^h_{init}.
\end{equation}
In practice, since the coarse prediction model captures most of the features in $Y^h$, the resulting distribution of $\Delta Y^h$ typically exhibits small magnitude in both mean and variance. Nevertheless, $\Delta Y^h$ often contains critical fine-grained motion corrections that can significantly improve prediction quality, particularly for the overly diverse samples generated by the coarse prediction model. The small variance in $\Delta Y^h$ makes Flow Matching struggle to learn a meaningful velocity field. In this study, we address this issue by scaling $\Delta Y^h$ with a scalar $\alpha$ to increase its variance. Specifically, we first estimate the variance of the prediction residuals from the training dataset and then choose $\alpha$ to scale the residuals such that the resulting variance in the training set approximately equals to 1. Therefore, the loss function in our study becomes

\begin{equation}
    \mathcal{L}_{\text{refiner}}(\theta) = \mathbb{E}_{t, Y_0, \Delta Y^h} [ \| u_t^\theta (Y_t) - (\alpha \Delta Y^h - Y_0) \|^2 ]
\end{equation}
\begin{equation}
    Y_t = t \alpha \Delta Y^h + (1-t)Y_0
\end{equation}
where $t \sim \mathcal{U}[0,1]$ and $Y_0 \sim \mathcal{N}(0,I)$.

\subsection{Interaction-aware Refiner}

Collaborating in close proximity naturally fosters rich interactions between agents. However, previous motion prediction studies for HRC have largely overlooked this aspect and treat human motion in isolation, ignoring interactions between agents. In this work, we account for the influence of robot motion on human movements by conditioning the refiner module on observed robot motion and encouraging the model to learn the interaction between human and robot. In addition to the observed robot motion, the refiner is also conditioned on the observed human motion and the initial prediction, enabling it to capture sufficient information and learn the corresponding residuals. Inspired by SE-Transformer, which effectively learns token-wise interconnections \cite{tian2024transfusion}, we adapt SE-Transformer block to strengthen residual learning in this work. Detailed refiner architecture can be seen in Figure \ref{architecture}b.

Specifically, for human-related conditions, such as observed human motion and coarse human motion prediction, we follow the same conditioning strategy as \cite{tian2024transfusion}, which adds the conditioning information as additional input tokens. To get the condition tokens, we first pad the observed motion sequence $X_o^h$ to the same length as the complete sequence $X^h$ by repeating the last observed frame, and then apply DCT to get the corresponding frequency coefficients i.e.,
\begin{equation}
    Y_o^h = DCT(Padding(X_o^h)).
\end{equation}
Then, we flatten both $Y_o^h$ and $\hat{Y}_{init}^h$ and pass them through different embedding layers to compress the information into a single token, 
\begin{equation}
    C^h = Embed(Flat(Y_o^h)).
\end{equation}
\begin{equation}
    C_{init} = Embed(Flat(\hat{Y}_{init}^h)).
\end{equation}
We also embed the Flow Matching step time $t$  and add it to the human-related condition tokens to provide the network with the corresponding time information for Flow Matching. Combined with positional embeddings, all tokens are passed through several SE-Transformer blocks, with concatenation-based residual connections linking the shallow and deeper layers. By leveraging the attention mechanism, the network can learn the relevance among tokens. Consequently, the model avoids relying on computationally expensive blocks, such as cross-attention, to handle human-related conditions, making the network smaller and faster.

However, such an in-context conditioning mechanism is implicit and the model is not explicitly constrained to use the condition tokens. In the extreme case, the model might even ignore these tokens, assigning them low attention scores that translate into negligible weights after the softmax operation in the self-attention mechanism. To encourage the network to learn human-robot interactions, we adopt an explicit conditioning strategy, the adaptive layer normalization (AdaLN) \cite{peebles2023scalable}, for the observed robot motion, as shown in Figure \ref{architecture}c. Specifically, we compute the robot condition $C^r$ as
\begin{equation}
    C^r = Embed(Flat(DCT(Padding(X_o^r)))),
\end{equation}
and use it in each interaction-aware SE-Transformer block to derive the modulation parameters: the dimension-wise scale $\gamma$ and shift $\beta$ applied after each layer normalization, and the gate $\alpha$ applied before the residual connection. Since these parameters are functions of $C^r$, the robot-conditioned information is consistently injected into the latent variables across multiple layers. This explicit conditioning mechanism ensures that the network integrates the robot’s motion into its internal representations, thereby enforcing the modeling of human-robot interactions.

\subsection{PrediFlow}

To formulate a prediction-refinement framework, we use a pretrained stochastic motion prediction model, SwiftDiff \cite{tian2024bayesian}, as the prediction module, since it generates predictions in a single denoising step, demonstrating time efficiency for real-time human motion prediction. The architecture is illustrated in Figure \ref{architecture}a. Specifically, the human motion history condition and noise input tokens are first embedded through their respective embedding layers and then concatenated. The combined representation is subsequently processed by multiple spatial-temporal MLP layers, mixing information across both dimensions. The output of the final MLP layer is then passed through a linear projection layer and a denoising step to produce coarse motion predictions in a single inference step. We then employ the proposed flow matching-based, interaction-aware residual learning model as a refiner. The parameters of the prediction model are frozen during training, and only the parameters of the refiner are updated to learn the residuals. 

Given a sequence of observations, the predictor first generates multiple coarse predictions. For each preliminary prediction, the refiner then produces multiple residuals in a single inference step. Since the residuals in the refinement stage contain significant uncertainty, stemming from both the data variation and the model's limited generalization, each residual contributes to improving prediction accuracy to different extents. Accordingly, these residuals can be aggregated in two ways to obtain the final refined prediction: (1) by computing the mean of multiple residuals and adding it to the initial prediction, which helps mitigate the impact of the variability through the estimated mean value, resulting in more consistent and reliable refinement outcomes; or (2) by adding individual residuals, which preserves the full stochasticity of the refinement stage and allows for a richer set of final predictions. More details on training and inference can be found in Algorithm \ref{alg1} and \ref{alg2}. Note that for inference, multiple predictions and residuals can be generated in batch.

\begin{algorithm}[H]
	\renewcommand{\algorithmicrequire}{\textbf{Input:}}
	\renewcommand{\algorithmicensure}{\textbf{Output:}}
	\caption{Training of PrediFlow}
	\label{alg1}
	\begin{algorithmic}[1]
            \REQUIRE Pretrained stochastic motion predictor $SwiftDiff(\cdot)$, Refiner $u^\theta$, HRC dataset $(X^h, X^r)$, Scaling parameter $\alpha$, Maximum training epoch $E_{max}$.
		\FOR{$i = 0,1,\cdots,E_{max}$}
            \STATE \hspace{1em} \textcolor{gray}{\# Sample data from HRC dataset}
		\STATE $X^h, X^r \sim p(X^h, X^r), \quad X^h, X^r \to X_o^h, X_o^r$
            \STATE $C^h = Embed(Flat(DCT(Padding(X_o^h))))$
            \STATE $C^r = Embed(Flat(DCT(Padding(X_o^r))))$
            \STATE $Y^h = DCT(X^h)$
            \STATE \hspace{1em} \textcolor{gray}{\# Prediction model generate one coarse prediction}
            \STATE $\hat{Y}_{init}^h = SwiftDiff(C^h)$
            \STATE $C_{init} = Embed(Flat(\hat{Y}_{init}^h))$
            \STATE \hspace{1em} \textcolor{gray}{\# Calculate the ground truth residual}
            \STATE $\Delta Y^h = Y^h - \hat{Y}^h_{init}$
            \STATE \hspace{1em} \textcolor{gray}{\# Train the refiner}
            \STATE $t \sim \mathcal{U}[0,1], \quad Y_0 \sim \mathcal{N}(0,I)$
            \STATE $Y_t = t \alpha \Delta Y^h + (1-t)Y_0$
            \STATE Take gradient descent step on \\ $ \nabla _ {\theta} \| u^\theta (Y_t, C_{init}, C^h, C^r, t) - (\alpha \Delta Y^h - Y_0) \|^2 $
            \ENDFOR
		\ENSURE  Trained refiner network $u^\theta$
	\end{algorithmic}
\end{algorithm}

\begin{algorithm}[H]
	\renewcommand{\algorithmicrequire}{\textbf{Input:}}
	\renewcommand{\algorithmicensure}{\textbf{Output:}}
	\caption{Inference of PrediFlow}
	\label{alg2}
	\begin{algorithmic}[1]
            \REQUIRE Stochastic motion predictor $SwiftDiff(\cdot)$, Refiner $u^\theta$, Observed motion $X_o^h$ and $X_o^r$, Scaling parameter $\alpha$, Aggregation option $agg \in \{\text{mean}, \text{all}\}$.
            \STATE \hspace{1em} \textcolor{gray}{\# Process the observation}
            \STATE $C^h = Embed(Flat(DCT(Padding(X_o^h))))$
            \STATE $C^r = Embed(Flat(DCT(Padding(X_o^r))))$
            \STATE \textcolor{gray}{\# Predict $N$ future motions}
            \FOR{$i = 0,1,\cdots,N$}
            \STATE $\hat{Y}_{init}^{h,i} = SwiftDiff(C^h)$
            \STATE $C_{init}^i = Embed(Flat(\hat{Y}_{init}^{h,i})))$
            \STATE \textcolor{gray}{\# Generate $M$ residuals for each coarse prediction}
            \FOR{$j = 0,1,\cdots,M$}
            \STATE $Y_0^j \sim \mathcal{N}(0,I)$
            \STATE $d^j = u^\theta (Y_0^j, C_{init}^i, C^h, C^r, 0)$
            \STATE \textcolor{gray}{\# One-step integral}
            \STATE $\Delta \hat{Y}^{h,ij} = (Y_0^j + d^j) / \alpha$
            \ENDFOR
            \IF {$agg = \text{mean}$}
            \STATE \textcolor{gray}{\# Calculate the mean of $M$ residuals}
            \STATE $\Delta \hat{Y}^{h,i} = mean(\{\Delta\hat{Y}^{h,ij}\}_{j=1,\cdots,M})$
            \STATE \textcolor{gray}{\# Refine the coarse prediction and apply IDCT}
            \STATE $\hat{X}^{h,i} = IDCT(\hat{Y}_{init}^{h,i} + \Delta \hat{Y}^{h,i})$
            \ELSIF{$agg = \text{all}$}
            \STATE \textcolor{gray}{\# Use all residuals  to preserve stochasticity}
            \STATE $\hat{X}^{h,ij} = IDCT(\hat{Y}_{init}^{h,i} + \Delta \hat{Y}^{h,ij}), \forall j = 1,\cdots,M$
            \ENDIF
            \ENDFOR
		\ENSURE $\{\hat{X}^{h,i}\}_{i = 1, \cdots, N}$ or $\{\hat{X}^{h,ij}\}_{i=1,\cdots,N;j=1,\cdots,M}$
	\end{algorithmic}
\end{algorithm}

\begin{table*}[htbp]
\caption{Percentage of Improvement with the Refiner under Different Evaluation Strategies on the HRC Dataset}
\centering
\resizebox{2.0\columnwidth}{!}{%
\begin{threeparttable}
\begin{tabular}{ccccc}
\toprule
& \multicolumn{4}{c}{HRC Dataset (Best-of-many / Median-of-many / Worst-of-many)} \\ \cmidrule{2-5} 
Model & ADE-B/M/W (m) & FDE-B/M/W (m) & MMADE-B/M/W (m) & MMFDE-B/M/W (m) \\ \midrule
SwiftDiff & 0.163 / 0.248 / 0.506 & 0.219 / 0.406 / 0.938 & 0.228 / 0.322 / 0.574 & 0.270 / 0.464 / 0.988 \\
PrediFlow & 0.159 / 0.221 / 0.374 & 0.232 / 0.368 / 0.664 & 0.230 / 0.299 / 0.446 & 0.288 / 0.428 / 0.717 \\ \midrule
Percentage of Improvement & \textbf{2.454\%} / \textbf{10.887\%} / \textbf{26.087\%} & -5.936\% / \textbf{9.360\%} / \textbf{29.211\%} & -0.877 \% / \textbf{7.143\%} / \textbf{22.300\%} & -6.667\% / \textbf{7.759\%} / \textbf{27.429\%} \\ \bottomrule
\end{tabular}
\begin{tablenotes}
     \item[*] Improvements are highlighted in bold. SwiftDiff refers to the model that balances inference time and accuracy in \cite{tian2024bayesian}. Quantitative results for PrediFlow are reported using the mean aggregation strategy for robustness.
\end{tablenotes}
\label{hrc_result}
\end{threeparttable}
}
\end{table*}

\begin{table*}[htbp]
\caption{Comparison with other predictors and refiners}
\centering
\resizebox{2.0\columnwidth}{!}{%
\begin{threeparttable}
\begin{tabular}{cccccc}
\toprule
& \multicolumn{5}{c}{HRC Dataset (Best-of-many / Median-of-many / Worst-of-many)} \\ \cmidrule{2-6} 
Model & Inference Time (sec) & ADE-B/M/W (m) & FDE-B/M/W (m) & MMADE-B/M/W (m) & MMFDE-B/M/W (m) \\ \midrule
DLow  & 0.0752 & 0.232 / 0.730 / 1.516 & 0.294 / 1.091 / 2.238 & 0.252 / 0.741 / 1.524 & 0.302 / 1.091 / 2.239 \\
HumanMAC  & 1.338 & \textbf{0.141} / 0.249 / 0.557 & \textbf{0.190} / 0.422 / 1.069 & \textbf{0.215} / 0.332 / 0.621 & \textbf{0.247} / 0.488 / 1.108 \\
TransFusion & 1.194  & 0.145 / 0.252 / 0.556 & 0.198 / 0.423 / 1.036 & 0.217 / 0.330 / 0.619 & 0.253 / 0.485 / 1.074 \\
SwiftDiff  &  \textbf{0.00618} & 0.163 / 0.248 / 0.506 & 0.219 / 0.406 / 0.938 & 0.228 / 0.322 / 0.574 & 0.270 / 0.464 / 0.988 \\ \midrule
SwiftDiff + Stoch. Refiner & 0.0105  & 0.162 / 0.225 / 0.379  & 0.238 / 0.377 / 0.686 & 0.230 / 0.300 / 0.447 & 0.287 / 0.432 / 0.733 \\
SwiftDiff + Det. Interaction-aware Refiner & 0.0115 & 0.168 / 0.231 / 0.392 & 0.239 / 0.376 / 0.680 & 0.232 / 0.302 / 0.458 & 0.288 / 0.430 0.732 \\
SwiftDiff + GCN & 0.0103 & 0.172 / 0.236 / 0.385 & 0.252 / 0.393 / 0.700 & 0.236 / 0.306 / 0.450 & 0.296 / 0.442 / 0.743 \\ \midrule
PrediFlow & 0.0120 & 0.159 / \textbf{0.221} / \textbf{0.374} & 0.232 / \textbf{0.368} / \textbf{0.664} & 0.230 / \textbf{0.299} / \textbf{0.446} & 0.288 / \textbf{0.428} / \textbf{0.717} \\ 
\bottomrule
\end{tabular}
\begin{tablenotes}
     \item[*] Best results for each metric are highlighted in bold. Stoch. refers to stochastic and Det. refers to deterministic. GCN-based refiner is taken from \cite{wei2023human}. As data streams at 60 Hz, the time budget for real-time prediction is 0.0167 seconds. Baselines are retrained using their official implementations and carefully tuned on the HRC dataset for optimal results. Quantitative results for PrediFlow are reported using the mean aggregation strategy for robustness.
\end{tablenotes}
\label{comparison}
\end{threeparttable}
}
\end{table*}

\begin{table*}[htbp]
\caption{Comparison of Residual Aggregation Strategies on the HRC Dataset}
\centering
\resizebox{2.0\columnwidth}{!}{%
\begin{threeparttable}
\begin{tabular}{cccccc}
\toprule
&& \multicolumn{4}{c}{HRC Dataset (Best-of-many / Median-of-many / Worst-of-many)} \\ \cmidrule{3-6} 
Model & Aggregation Strategy & ADE-B/M/W (m) & FDE-B/M/W (m) & MMADE-B/M/W (m) & MMFDE-B/M/W (m) \\ \midrule
SwiftDiff & N/A & 0.163 / 0.248 / 0.506 & 0.219 / 0.406 / 0.938 & 0.228 / 0.322 / 0.574 & 0.270 / 0.464 / 0.988 \\ \midrule
PrediFlow & Adding Individual Residuals for Full Stochasticity & \textbf{0.150} / 0.223 / 0.416 & \textbf{0.210} / 0.371 / 0.753 & \textbf{0.220} / 0.301 / 0.488 & \textbf{0.264} / 0.431 / 0.805 \\
PrediFlow & Adding Mean of Residuals for Robustness & 0.159 / \textbf{0.221} / \textbf{0.374} & 0.232 / \textbf{0.368} / \textbf{0.664} & 0.230 / \textbf{0.299} / \textbf{0.446} & 0.288 / \textbf{0.428} / \textbf{0.717} \\ \bottomrule
\end{tabular}
\begin{tablenotes}
     \item[*] Best results for each metric are highlighted in bold. For each coarse prediction, 10 residuals are generated in parallel. SwiftDiff is a prediction-only model, which serves as the predictor in our prediction-refinement framework, PrediFlow.
\end{tablenotes}
\label{hrc_strategy}
\end{threeparttable}
}
\end{table*}

\section{Experiments}

\subsection{Experimental Setup}

\textbf{HRC Datasets:} While many benchmark datasets exist for regular human motion, such as walking, running, and other daily activities, large-scale datasets focusing on human-robot close collaboration in manufacturing, with rich interaction between human and robot, remain scarce. In this work, we use an HRC desktop disassembly dataset \cite{tian2024bayesian} that simulates real-world end-of-life product disassembly and recycling to evaluate our interaction-aware prediction-refinement framework, PrediFlow. In the experiment, one human and one robot collaborate across a shared workstation to fully disassemble a used desktop. Dexterous tasks unsuitable for the robot’s two-finger gripper are handled by the human, while the robot assists by removing components released by the human (e.g., memory cards, hard drives), handing over tools, and placing disassembled parts into a recycling bin. The interactions between the human and robot occur in disassembly experiments when the human operator actively avoids collisions with the robot, when components are handed over between the human and the robot, or when the human assists the robot with specific tasks.

The dataset contains over two and a half hours of motion data and includes 32 complete disassembly trials, comprising a total of 0.6 million data frames captured at 60 Hz. The disassembly sequence for each trial is completely different to increase the data variation. We follow the train-validation-test split from \cite{tian2024bayesian}, using 24 trials for training, 4 for validation, and 4 for testing. We use 0.5 seconds (30 frames) of data to predict the next 2 seconds (120 frames) of human motion, following \cite{tian2024bayesian}. We set the multi-modal threshold \cite{yuan2020dlow}, which is a criterion for determining whether two observed motions are similar, to 0.2 in this work. At this threshold, 6.22\% of validation and 6.64\% of test motions are unique, while around 40\% of motions in both sets have fewer than 10 similar samples, which is comparable to the ratios in \cite{yuan2020dlow}.

\textbf{Evaluation Metrics:} We use the same evaluation strategy as in \cite{tian2024bayesian}. Specifically, Five metrics are considered: Inference Time for determining real-time prediction capability, Average Displacement Error (ADE) which calculates the average L2 distance over all time steps between the ground truth and the predicted sample, Final Displacement Error (FDE) which calculates the L2 distance in the last time frame between the ground truth and the predicted sample, and the multi-modal versions of ADE and FDE (MMADE \& MMFDE) for measuring prediction accuracy. Moreover, for each accuracy metric, three cases are considered: Best-of-many, Median-of-many, and Worst-of-many, to provide a comprehensive evaluation of the overall prediction quality.

\textbf{Baselines:} Considering that SwiftDiff is the only stochastic model for 3D human motion prediction that demonstrates both good prediction accuracy and real-time prediction capacity \cite{tian2024bayesian}, we first compare our method with several SwiftDiff-refiner combinations. Specifically, we compared PrediFlow with a flow-based refiner using the same network architecture but without considering human-robot interaction, referred to as the Stochastic Refiner, and a deterministic refiner with a similar network that considers the interaction as well, referred to as the Deterministic Interaction-aware Refiner, as well as the GCN-based Refiner from \cite{wei2023human}. To demonstrate that other existing stochastic predictors suffer from significant inference latency, we include three additional predictors in the baselines: HumanMAC \cite{chen2023humanmac} and TransFusion \cite{tian2024transfusion}, which are two advanced diffusion-based predictors that achieve state-of-the-art prediction accuracy, as well as DLow \cite{yuan2020dlow}, a VAE-based predictor with faster inference speed compared to diffusion models.

\textbf{Implementation Details:} The parameters of the pretrained SwiftDiff are frozen, and we train only the refiner in this study. We set the training epoch to 1,000 with an initial learning rate of $2.5 \times 10^{-4}$. In each epoch, we randonmly sample 50,000 data points from the training dataset to train the model, and the batch size is set to 64. Training is warmed up for 100 epochs before applying the cosine annealing learning rate decay strategy. We use a 7-layer interaction-aware SE-Transformer with a latent dimension of 512 in the refiner. For each prediction, 10 residuals are generated for evaluation. Baselines are fine-tuned for optimal performance. Inference time is tested using an NVIDIA RTX 4080, which is more commonly available. All other experiments, including training and evaluation, are conducted on a single NVIDIA A100 GPU.

\subsection{Effectiveness of Refiner Network}

\begin{figure*}
    \centering
    \includegraphics[width=1.95\columnwidth]{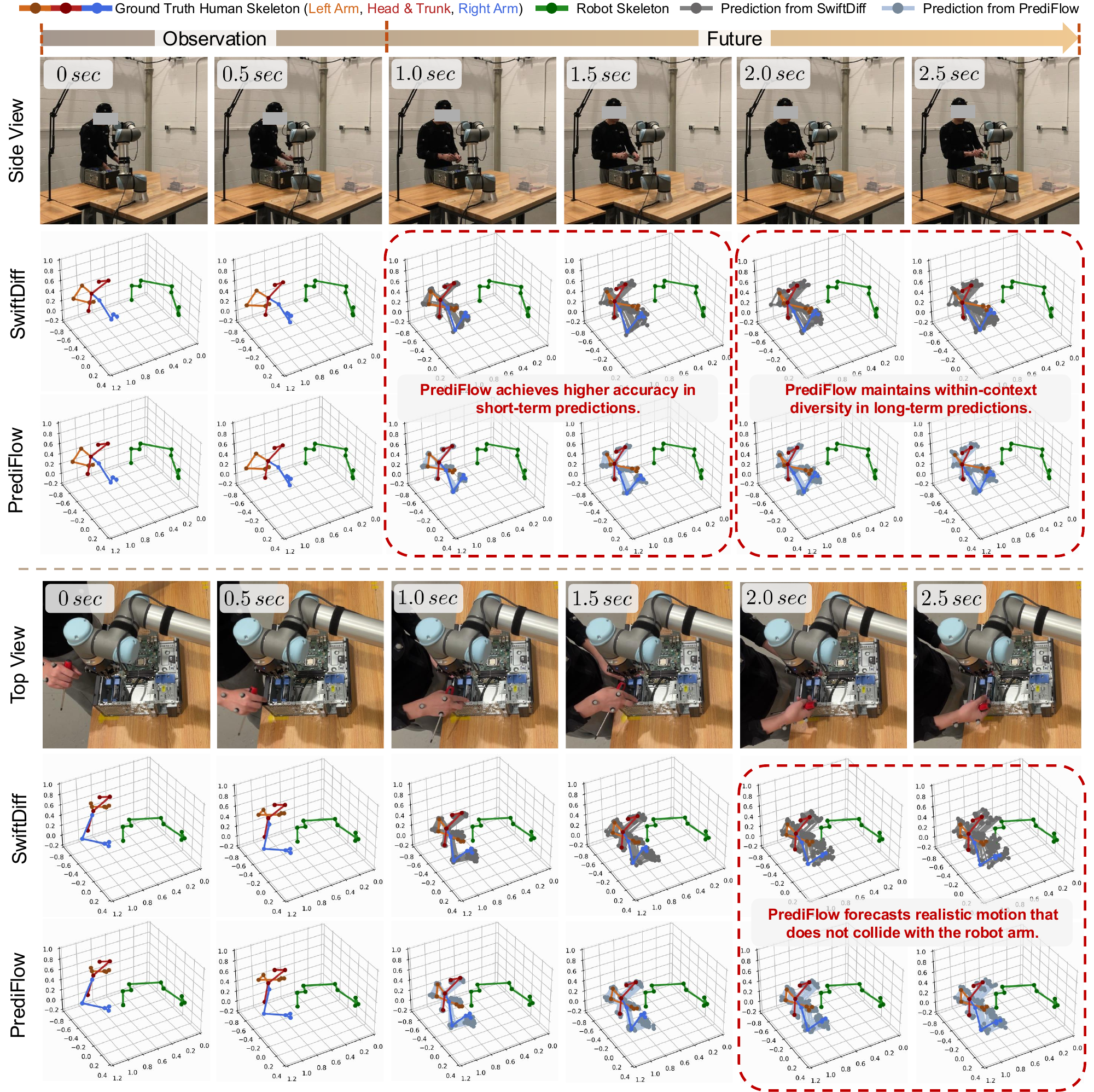}
    \caption{Prediction Visualization. The first row of each case presents the top or side view of the experiment, while the last two rows display the animation of prediction results from SwiftDiff and PrediFlow, respectively. Case 1 (top): The human is disassembling an expansion card from the desktop case and handing it to the robot. In this case, the prediction-refinement framework makes short-term predictions more consistent and accurate while still capturing within-context uncertainty in long-term predictions. Case 2 (bottom): The human assists robot manipulator to disassemble hard drive. By considering human-robot interaction, the prediction-refinement framework generates realistic, collision-free motions, whereas SwiftDiff predicts motions that result in collisions.}
    \label{case_1}
\end{figure*}

\begin{figure}
    \begin{center}
    \includegraphics[width=0.95\columnwidth]{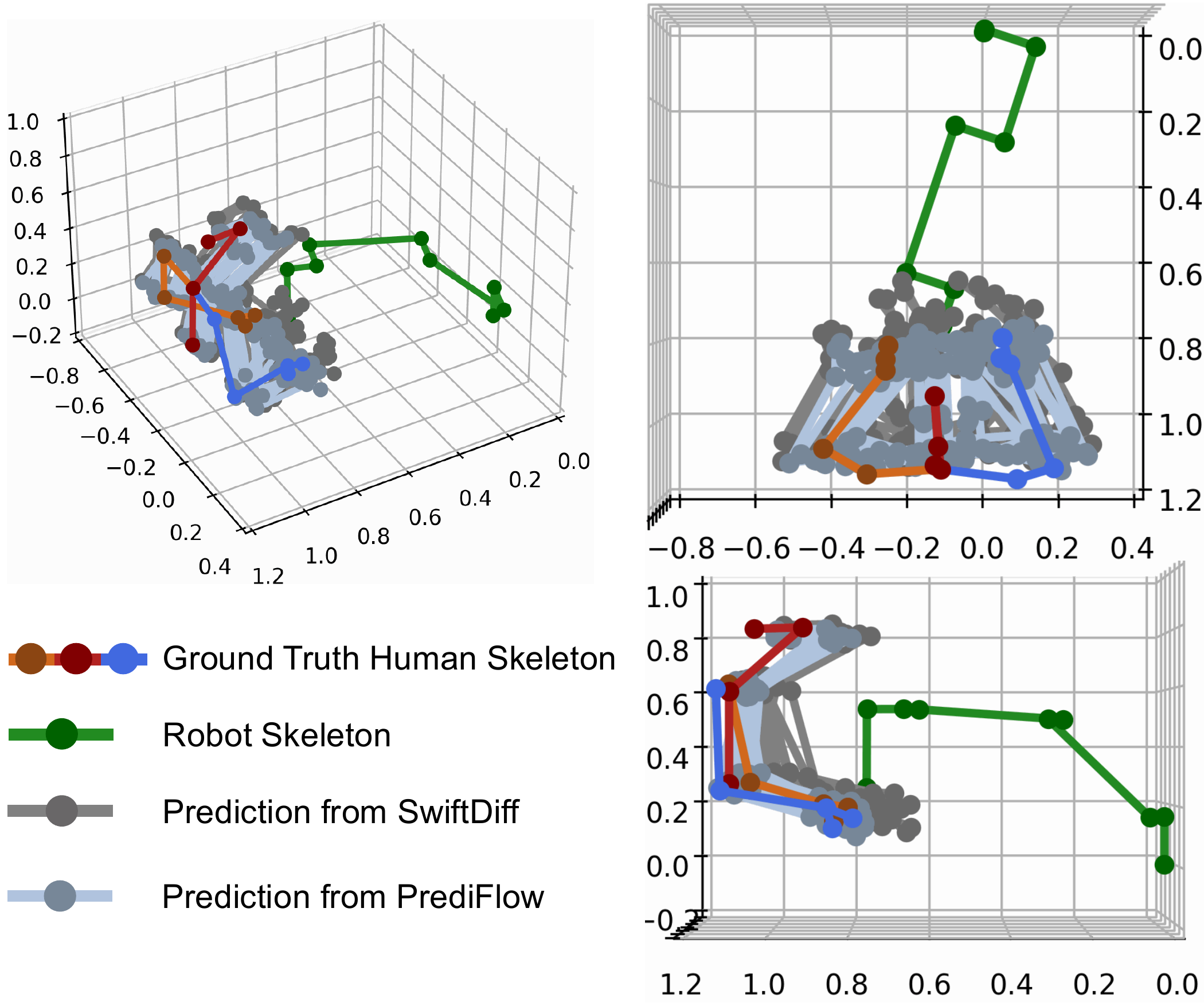}
    \caption{Different views of the last-frame prediction in Experiment Case 2.}
    \label{case_2_zoom}
    \end{center}
\end{figure}

As stated earlier, the residuals can be aggregated with the initial predictions in two ways to obtain the final results: by adding individual residuals to account for full stochasticity, or by adding their mean to provide a more robust and consistent improvement. We first compare our model with existing works using the mean strategy for robustness in Table \ref{hrc_result} and Table \ref{comparison}, and then demonstrate how variation in the residuals affects the refinement later in Table \ref{hrc_strategy}.

To showcase the effectiveness of our prediction-refinement framework, we provide the quantitative results over the entire test dataset, comparing our method with prediction-only baseline, SwiftDiff, in Table \ref{hrc_result}. PrediFlow consistently improves prediction accuracy across multiple metrics, achieving a 7\% to 10\% error reduction in Median-of-many case and a 22\% to 30\% reduction in the Worst-of-many case. Though there is an error increase in the Best-of-many FDE, MMADE, and MMFDE, the degradation is subtle and negligible compared to the overall improvement. Specifically, the error increases by 13 millimeters in FDE-Best, 2 millimeters in MMADE-Best, and 18 millimeters in MMFDE-Best. However, in the median and worst cases, the error significantly decreases by approximately 3 centimeters and 20 centimeters, respectively. Such consistent improvement in the results of Median-of-many and Worst-of-many cases, but not the Best-of-many case is because the refiner, which is designed to learn fine-grained adjustments, is particularly effective at addressing prediction errors in suboptimal samples which initially deviate more from the ground truth. In contrast, the Best-of-many strategy selects the prediction that already aligns most closely with the ground truth. Since this prediction is already highly accurate, there is little room for improvement.
By adding the mean of several generated residuals, the model may have limited improvement or even slightly negative effects in the Best-of-many case.

Moreover, despite introducing additional refinement module to correct the human motion prediction, which make the predictions more consistent and reduce their diversity, the multi-modal accuracies are even better than those of previous approach, indicating that we still maintain meaningful diversity while improving robustness and overall prediction accuracy by eliminating unrealistic and extreme predictions. Our method, in this case, can make downstream applications, such as robot motion planning, less conservative, as the predicted future motion occupies less shared space, thereby increasing the planning success rate and collaboration efficiency while maintaining the safety of human collaborators.

We present the prediction visualization in Figure \ref{case_1}. In the first case (Figure \ref{case_1} top), where the human is removing the expansion card and handing it to the robot manipulator, the PrediFlow model captures the interaction between the human and robot, i.e., the robot is requesting disassembled components from the human to recycle them, making it relatively confident in short-term prediction and resulting in more consistent and accurate predictions compared to SwiftDiff. As uncertainty accumulates over time, PrediFlow becomes more diverse in long-term prediction, while remaining contextually relevant. In the second case (Figure \ref{case_1} bottom), where the human assists the robot manipulator in hard drive disassembly, by considering human-robot interaction, PrediFlow generates collision-free predictions. In contrast, SwiftDiff predicts motions that result in collisions, which can be clearly observed in Fig. \ref{case_2_zoom}. It is worth to note that we do not add explicit collision constraints in the loss function, because during HRC, the human does not always intend to avoid the robot. In fact, in some cases, such as during handover tasks or collaborative disassemble the same component, the human intentionally stays close to the robot. Instead of introducing hard collision avoidance constraints, our proposed refiner effectively learns the interaction between the human and the robot, and predicts appropriate human behavior during collaboration. We also conduct an online prediction experiment to demonstrate the real-time prediction capability, and the visualization is provided in the supplementary video\footnote{The supplementary video can be accessed via the link: \url{https://drive.google.com/file/d/1wV6YYrSEUc1fwO2qRxapiFcWZKkM-LMY/view}}.

\subsection{Comparison with Baseline Predictors and Refiners}

To compare our method with existing works, we first demonstrate that other predictors require significantly longer inference times compared to SwiftDiff, which is chosen as the initial predictor in this work, by comparing the first four rows in Table \ref{comparison}. The inference time is measured by providing one single sequence of observed motion and generating 50 predictions in parallel, with the results averaged over 20 runs. Although the VAE-based DLow model is much faster than the other two diffusion-based baselines, it still exceeds the time budget and yields lower prediction accuracy than the others.

Since the contribution of our refiner is twofold: first, we use the Flow Matching framework to capture uncertainty in residual learning, and second, we explicitly introduce the observed robot motion into the refiner network, making it an interaction-aware refiner, to evaluate its effectiveness, we compare PrediFlow with other refiner combinations in Table \ref{comparison}. By comparing the fifth row with the last row, we can conclude that including the robot motion further improves accuracy. The error can be reduced by up to approximately 2 centimeters, as shown in the FDE-W and MMFDE-W metrics. Similarly, by comparing the sixth row and the last row, all accuracy metrics improve with the stochastic model. To further compare with an existing refinement work \cite{wei2023human}, we use the open-source code to fine-tune the GCN network on HRC dataset, and the results are shown in the seventh row of Table \ref{comparison}. PrediFlow consistently outperforms the GCN-based method in terms of prediction accuracy, demonstrating the benefits of our novel framework.

Furthermore, we compare the inference times of different refiner combinations with SwiftDiff. Although the inference speed of PrediFlow is slightly slower than the combination of SwiftDiff and GCN-based refiner, considering that the data streams at 60 Hz, all the frameworks have the capability for real-time prediction, and can be applied in the real-world robot perception and decision making tasks. As long as real-time prediction constraints are met, the model that delivers higher prediction accuracy should be preferred.

\subsection{Comparison of residual aggregation strategies}

To evaluate the effectiveness of different residual aggregation strategies, we compare the predictor-only model, SwiftDiff, with PrediFlow using two approaches: (1) adding individual residuals and (2) adding the mean of generated residuals, as shown in Table \ref{hrc_strategy}. For each coarse prediction, we generate ten residuals through our proposed refiner to account for the uncertainty in the refinement stage. As noted earlier, each residual contributes differently to improving prediction accuracy. Adding all residuals individually to the coarse prediction preserves the full stochasticity of the refinement process, allowing the model to capture more diverse plausible outcomes. In contrast, estimating the mean of the residuals and adding it to the coarse prediction reduces the impact of such variation, thereby enhancing robustness and producing more consistent results. This distinction is evident when comparing the second and third rows of Table \ref{hrc_strategy}: the range of displacement errors, i.e., the difference between the Best-of-many and the Worst-of-many displacement error, for adding individual residuals is consistently larger than that of the mean residual strategy across all four evaluation metrics. Specifically, the Best-of-many results demonstrate lower errors when individual residuals are added, highlighting the advantage of preserving stochasticity. However, this approach also leads to higher errors in the Worst-of-many case. Notably, although the Worst-of-many errors for adding individual residuals are larger than those for the mean residual strategy, they remain smaller than those of the predictor-only model. This indicates that even residuals with a smaller individual impact can still contribute to improving prediction accuracy, rather than degrading it.

Both strategies are practical for real-world applications. Depending on the user preference, one can prioritize preserving more diverse potential motions by adding individual residuals, or favor consistency and reduced variability by using mean residual aggregation. This flexibility allows our framework to adapt to a wide range of human-robot collaborative scenarios.

\section{Conclusion}

This paper presents PrediFlow, a prediction-refinement framework for real-time 3D human motion prediction in HRC disassembly. We leverage a pretrained stochastic motion prediction network as the prediction module, and propose a novel flow-based interaction-aware prediction refiner. 
Results show that our proposed coarse-to-fine strategy, which captures uncertainty and models the effect of robot motion on human behavior, significantly improves prediction accuracy while preserving meaningful diversity. Although there is an additional refinement block, PrediFlow can still operate within the real-time inference budget, demonstrating its practical applicability.
We acknowledge several limitations in our work.
First, since the primary focus of this work is to develop an advanced interaction-aware human motion prediction module that helps robots better understand human behavior during collaboration, the current human motion predictions do not yet directly influence robot planning decisions. Integrating the prediction module into robot control and planning is an important direction for future research. We believe this work represents a foundational step toward enabling responsive and adaptive robot planning that enhances efficiency and safety in HRC. Additionally, due to the lack of large-scale, interaction-rich HRC datasets in manufacturing, the proposed method is trained only on a customized HRC disassembly dataset. Collecting a large-scale HRC dataset for manufacturing represents a promising direction for future research.

\bibliographystyle{IEEEtran}
\bibliography{ref}{}

\end{document}